\documentclass{article}
\usepackage[utf8]{inputenc}
\usepackage{authblk}
\usepackage{setspace}
\usepackage[margin=1.25in]{geometry}
\usepackage{graphicx}
\graphicspath{ {./figures/} }
\usepackage{subcaption}
\usepackage{amsmath}
\usepackage[T1]{fontenc}
\usepackage{float}
\usepackage{amsfonts}
\usepackage{amssymb}
\usepackage[T1]{fontenc}
\usepackage{lmodern,tikz,lipsum}
\usepackage{fancyhdr}
\usepackage{changepage}
\usepackage{hyperref}
\usepackage{tocloft}
\usepackage{wallpaper}
\usepackage{lipsum} 
\usepackage{titlesec} 

\usepackage{algorithm}
\usepackage{algpseudocode}

\usepackage{amsthm}

\newtheorem{theorem}{Theorem}

\title{Super Gradient Descent: Global Optimization requires Global Gradient}

\author[1]{Seifeddine Achour}

\date{}

\begin{document}

\maketitle

\begin{abstract}
Global minimization is a fundamental challenge in optimization, especially in machine learning, where finding the global minimum of a function directly impacts model performance and convergence. This article introduces a novel optimization method that we called Super Gradient Descent, designed specifically for one-dimensional functions, guaranteeing convergence to the global minimum for any $k$-Lipschitz function defined on a closed interval $[a, b]$. Our approach addresses the limitations of traditional optimization algorithms, which often get trapped in local minima. In particular, we introduce the concept of global gradient which offers a robust solution for precise and well-guided global optimization. By focusing on the global minimization problem, this work bridges a critical gap in optimization theory, offering new insights and practical advancements in different optimization problems in particular Machine Learning problems like line search.

\end{abstract}


\section{Introduction}
Global optimization plays a critical role in addressing complex real-life challenges across various fields. In engineering, it is applied to structural design optimization, where minimizing weight or material use while ensuring durability is essential for cost-effective and safe construction. In financial services, portfolio optimization requires balancing risk and return by finding the global minimum or maximum in investment strategies. In logistics and transportation, global optimization is crucial for solving routing problems such as determining the shortest path or optimizing delivery routes which leads to significant cost savings and improved efficiency. Similarly, in energy systems, global optimization is key to managing and distributing power more efficiently, reducing operational costs, and optimizing renewable energy usage.\\\\
In machine learning, the need for global optimization is especially pronounced. The performance of models often depends on the ability to minimize complex, non-convex loss functions. While traditional methods like gradient descent are effective in many cases, they frequently encounter the problem of getting trapped in local minima, which can hinder the model’s overall performance. This is particularly relevant in tasks that require complex models where the optimization landscape is highly non-linear and fraught with local minima.\\\\
The primary contribution of this work is the introduction of a novel algorithm named Super Gradient
Descent. Unlike classical gradient descent, which collects only local information making it prone to local minima, the proposed method adapts the state’s change
decision based on a global detection of the function change to ensure consistent progress
towards the global minimum. We evaluate its performance on various one-dimensional
functions, demonstrating that it provides superior convergence behavior, particularly in
avoiding local minima and achieving the global optimum. This novel approach contributes
to overcoming the challenges of non-convex optimization, offering a more reliable method
for finding global solutions in machine learning.

\section{State of the Art}

Optimization algorithms are crucial in training machine learning models, as they guide the parameter updates in response to the loss function. Below, we discuss several prominent optimization algorithms and their mathematical formulations.

\subsection{Gradient Descent (GD)}  
Gradient Descent (GD) is a fundamental optimization algorithm used to minimize a given objective function by iteratively updating parameters in the direction of the steepest descent, which is indicated by the negative gradient of the function\cite{andrychowicz2016learning}.\\
The update rule for Gradient Descent is:
\[
x_{t+1} = x_t - \eta \nabla f(x_t)
\]
where:
\begin{itemize}
    \item \(x_t\) represents the vector of parameters at iteration \(t\),
    \item \(\eta\) is the learning rate, which controls the step size,
    \item \(\nabla f(x_t)\) is the gradient of the objective function at \(x_t\).
\end{itemize}

\textbf{Pros}: Intuitive algorithm and stable convergence for convex problems.

\textbf{Cons}:  Can get stuck in local minima, particularly in non-convex problems.

\subsection{AdaGrad}
AdaGrad is an adaptive learning rate optimization algorithm that adjusts the learning rate for each parameter based on the historical gradients and handles the sparse data well. Parameters with large gradients receive smaller updates, while those with smaller gradients are updated more significantly.\cite{duchi2011adaptive}

The AdaGrad update rule is:
\[
x_{t+1} = x_t - \frac{\eta}{\sqrt{G_t + \epsilon}} \nabla f(x_t)
\]
where:
\begin{itemize}
    \item \(G_t\) is the sum of the squares of the gradients up to time step \(t\),
    \item \(\epsilon\) is a small constant to avoid division by zero.
\end{itemize}

\textbf{Pros}: Automatically adjusts learning rates; works well with sparse data.  

\textbf{Cons}: The learning rate decreases too aggressively over time, leading to premature convergence.

\subsection{RMSprop}
RMSprop was developed to address the diminishing learning rate issue in AdaGrad by introducing an exponentially decaying average of squared gradients \cite{tieleman2012lecture}. This allows RMSprop to adapt the learning rate dynamically without the aggressive reduction that AdaGrad experiences.

The RMSprop update rule is:
\[
x_{t+1} = x_t - \frac{\eta}{\sqrt{E[\nabla f(x_t)^2] + \epsilon}} \nabla f(x_t)
\]
where:
\begin{itemize}
    \item \(E[\nabla f(x_t)^2]\) is the exponentially weighted average of the squared gradients,
    \item \(\epsilon\) is a small constant for numerical stability.
\end{itemize}

\textbf{Pros}: More robust to the diminishing learning rate issue.  

\textbf{Cons}: Sensitive to hyperparameters.

\subsection{Adam}
Adam combines the advantages of both AdaGrad and RMSprop by maintaining an exponentially decaying average of past gradients (first moment) and squared gradients (second moment) \cite{kingma2014adam}. This approach allows it to use adaptive learning rates, while also addressing the diminishing learning rate issue.

The update rule for Adam is:
\[
m_t = \beta_1 m_{t-1} + (1 - \beta_1) \nabla f(x_t)
\]
\[
v_t = \beta_2 v_{t-1} + (1 - \beta_2) \nabla f(x_t)^2
\]
\[
\hat{m}_t = \frac{m_t}{1 - \beta_1^t}, \quad \hat{v}_t = \frac{v_t}{1 - \beta_2^t}
\]
\[
x_{t+1} = x_t - \frac{\eta \hat{m}_t}{\sqrt{\hat{v}_t} + \epsilon}
\]
where:
\begin{itemize}
    \item \(m_t\) and \(v_t\) are the first and second moment estimates,
    \item \(\beta_1\) and \(\beta_2\) are hyperparameters that control the decay rates of these moments.
\end{itemize}

\textbf{Pros}: Efficient and well-suited for a wide variety of problems; adaptive learning rates.

\textbf{Cons}: Lacks formal convergence guarantees in non-convex settings.

\subsection{AdamW}
AdamW is a variant of the Adam optimizer that decouples weight decay from the gradient-based parameter updates \cite{loshchilov2017decoupled}. This leads to better generalization performance, especially in deep learning models, by applying weight decay directly to the parameters as follows:
\[
x_{t+1} = x_t - \frac{\eta \hat{m}_t}{\sqrt{\hat{v}_t} + \epsilon} - \eta \lambda x_t
\]
where:
\begin{itemize}
    \item \(\lambda\) is the weight decay coefficient.
\end{itemize}

\textbf{Pros}: Better generalization than Adam; reduces overfitting by using weight decay.  

\textbf{Cons}: Requires careful tuning of the weight decay parameter.

\subsection{Nesterov Accelerated Gradient (NAG)}
Nesterov Accelerated Gradient (NAG) introduces a look-ahead step that computes the gradient not at the current position, but at the point where the momentum term would take it \cite{dozat2016nesterov}. This improves the convergence rate by anticipating the trajectory of the updates as follows:
\[
v_{t+1} = \beta v_t + \eta \nabla f(x_t - \beta v_t)
\]
\[
x_{t+1} = x_t - v_{t+1}
\]
where:
\begin{itemize}
    \item \(v_t\) is the velocity (momentum term),
    \item \(\beta\) is the momentum coefficient.
\end{itemize}

\textbf{Pros}: Faster convergence than standard momentum-based methods in general.  

\textbf{Cons}: Slightly more complex to implement and tune.

\subsection{Heuristic Methods}
Heuristic optimization methods, such as Genetic Algorithms (GA) \cite{chelouah2000continuous}and Particle Swarm Optimization (PSO)\cite{marini2015particle}, are inspired by natural processes and can be useful for finding global optima. They are particularly advantageous in complex landscapes but come with their own limitations:\\
\textbf{Convergence Speed:} Heuristic methods can be slow to converge and often require a large number of evaluations to achieve satisfactory results.\\
\textbf{Noisy Landscapes:} They may struggle in noisy environments, where the fitness landscape can mislead the optimization process.\\
\textbf{Global Convergence:} While they are designed to search the global space, they do not guarantee convergence to the global minimum.\\\\
In summary, while state-of-the-art optimization algorithms provide powerful tools for training machine learning models, they suffer from limitations regarding local minima and global convergence guarantees. In the following section we will introduce our new algorithm Super Gradient Descent (SuGD) while proving its convergence to the global minimum of any k-lipschitz one-dimensional function defined on a domain $[a, b]$.

\section{Methodology}
In this section, we present our methodology for implementing the Super Gradient Descent Algorithm along with the mathematical proof of its guaranteed global convergence for any $k$-Lipschitz continuous function defined on the domain $[a, b]$.
\subsection{Super Gradient Descent Implementation}
In the general case, it is hard to explicitly calculate the gradient of a any function and in many cases we don't even know the function explicitly. The Finite Difference Method (FDM) is widely used to approximate derivatives using discrete points on a function's domain, with the \textbf{forward difference} formula $f'(x) \approx \frac{f(x+h) - f(x)}{h}$ providing a first-order accurate estimate for $f'(x)$ by evaluating the difference between $f(x+h)$ and $f(x)$ over a small step size $h$. Similarly, the \textbf{backward difference} approximation $f'(x) \approx \frac{f(x) - f(x-h)}{h}$ offers first-order accuracy using previous points, while the \textbf{central difference} scheme $f'(x) \approx \frac{f(x+h) - f(x-h)}{2h}$ yields second-order accuracy by averaging forward and backward differences \cite{shahab2024finite}.\\ Generally speaking, approximating the derivation means evaluating the function at two close points and calculating the ratio of the difference between the two outputs and the difference between the two points of evaluation. This allow us to calculate what we call the local gradient which is used in optimization to point to the local minimum through following the negative of its direction with an optimization step chosen small enough to ensure convergence. What we aim here to find a step $\alpha_\epsilon > 0$ where $\forall \alpha \in [0, \alpha_\epsilon]
$ the solution converges to the global minimum. Inspiring from the local gradient concept we introduce the concept of global gradient of a function $f$ defined on domain D on dimension $i$ as:  
\begin{equation}
F_i(x,y) = \frac{f(y) - f(x)}{y-x} 
\end{equation}\\
It evaluates the function at any two points instead of necessarily two neighbor points to capture its global variation. In that case, the local gradient can be a particular case where $\nabla f(x) = \lim_{h \to 0} F(x+h,x)$.\\\\
Based on the definition of the global gradient, we introduce our algorithm that guarantees convergence to a global minimum for any k-Lipschitz one-dimensional function defined on an interval $[a,b]$, that we called the Super Gradient Descent (SuGD) as follows:\\

\begin{algorithm}
\caption{Super Gradient Descent}
\label{alg1}
\begin{algorithmic}[Algorithm 1]
\State \textbf{Input:} $f$, domain $[a, b]$, tolerance $\eta > 0$ , optimization step $\alpha > 0$
\State \textbf{Initialize:} Choose initial points $x^{(1)}_{0} = a$, $x^{(2)}_{0} = b$
\While{$|x^{(2)}_{n} - x^{(1)}_{n}|(1+|F(x^{(2)}_{n},x^{(1)}_{n})|) > \eta$}
    \If {$f(x^{(2)}_{n})-f(x^{(1)}_{n}) < 0$}
        \State Set $x^{(1)}_{n+1} = x^{(1)}_{n} -\alpha (x^{(1)}_{n}-x^{(2)}_{n})(1-F(x^{(2)}_{n},x^{(1)}_{n}))$, $x^{(2)}_{n+1} = x^{(2)}_{n}$
    \Else
        \State Set $x^{(2)}_{n+1} = x^{(2)}_{n} -\alpha (x^{(2)}_{n}-x^{(1)}_{n})(1+F(x^{(2)}_{n},x^{(1)}_{n}))$, $x^{(1)}_{n+1} = x^{(1)}_{n}$
    \EndIf
\EndWhile
\State \textbf{Return:} Approximate minimum at $x_n = x^{(1)}_{n}$ 
\end{algorithmic}
\end{algorithm}
The algorithm leverages the global derivative information to ensure fast and well-guided convergence, with a global vision within the interval $[a, b]$ while mastering the neighborhood between points to ensure a local behavior that is similar to the behavior of local algorithms. Next, we will provide the theoretical basis behind this algorithm.\\

\subsection{Theoretical Convergence Guarantee}

Before introducing the proof of the algorithm's global convergence, we start by recalling that a function $f$ is $k$-Lipschitz if:
\[
|f(x) - f(y)| \leq k |x - y| \quad \forall x, y \in [a, b].
\]
This means that $\forall x, y \in [a, b], \quad |F(x,y)| = \frac{|f(x) - f(y)| }{|x - y|}\leq k$. In other words, Lipschitz continuity ensures that our Global Gradient is always finite which justifies its usefulness in the SuGD algorithm.\\

\begin{theorem}

Let $f$ be a 1-D, k-Lipschitz function defined on a bounded domain $[a,b]$ that admits one global minimum at $x^* \in [a,b]$. 
$\exists  \alpha_\epsilon>0, \forall \alpha \in [0, \alpha_\epsilon]$ an optimization step, where the Super Gradient Descent Algorithm \ref{alg1} converges to this global minimum.
More precisely, $\forall \epsilon>0 \text{small enough}, \exists  \alpha_\epsilon>0 \quad where \quad \forall \alpha \in [0, \alpha_\epsilon], \exists n_0>0 \text{ where } \forall n>n_0$ ,$|f(x_n) - f(x^*)|< \epsilon$
\end{theorem}

\begin{proof}
To ensure that $x^{(1)}_n$,$x^{(2)}_n$ converges to $x^*$ the proof of this theorem will consist of two main parts: \\

\textbf{Part 1:} 
\begin{equation}
\label{eq2}
\text{Prooving with recurrence that } \exists \alpha_\epsilon>0,\quad \forall n \in \mathbb{N} \quad , \forall \alpha \in [0, \alpha_\epsilon] \left\{
\begin{array}{l}
x^{(1)}_n < x^*\\
x^{(2)}_n > x^*
\end{array}
\right.
\end{equation}
Where $\epsilon$ is the approximation threshold.\\\\
 \begin{equation}
\text{For } n=0 \quad \left\{
\begin{array}{l}
x^{(1)}_0 = a < x^*\\
x^{(2)}_0 = b > x^*
\end{array}
\right.
\end{equation}

Assuming that the inequalities are true until an order n.\\
If the sequence is stationary the recurrence is trivial for it.
If not then
 \begin{equation}
\quad \left\{
\begin{array}{l}
x^{(1)}_{n+1} = x^{(1)}_{n} -\alpha (x^{(1)}_{n}-x^{(2)}_{n})(1-F(x^{(2)}_{n},x^{(1)}_{n})), \text{ 
 when } f(x^{(2)}_{n})-f(x^{(1)}_{n})  < 0\\
x^{(2)}_{n+1} = x^{(2)}_{n} -\alpha (x^{(2)}_{n}-x^{(1)}_{n})(1+F(x^{(2)}_{n},x^{(1)}_{n})), \text{ 
 when } f(x^{(2)}_{n})-f(x^{(1)}_{n}) \geq 0
\end{array}
\right.
\end{equation}
Let's choose $\alpha_\epsilon = \frac{\epsilon}{(b-a)(1+k)k}, \quad 0 \leq \alpha \leq \alpha_\epsilon$\\\\
We have $|(x^{(1)}_{n}-x^{(2)}_{n})(1-F(x^{(2)}_{n},x^{(1)}_{n}))| = |x^{(1)}_{n}-x^{(2)}_{n} + f(x^{(2)}_{n})-f(x^{(1)}_{n})|$\\\\
And $|x^{(1)}_{n}-x^{(2)}_{n} + f(x^{(2)}_{n})-f(x^{(1)}_{n})| \leq|x^{(1)}_{n}-x^{(2)}_{n}| + |f(x^{(1)}_{n})-f(x^{(2)}_{n})|$\\\\
And from k-Lip condition $|x^{(1)}_{n}-x^{(2)}_{n}| + |f(x^{(1)}_{n})-f(x^{(2)}_{n})|\leq (b-a)(1+k)$\\\\
Therefore $|(x^{(1)}_{n}-x^{(2)}_{n})(1-F(x^{(2)}_{n},x^{(1)}_{n}))|\leq (b-a)(1+k)$\\\\
Means $\alpha|(x^{(1)}_{n}-x^{(2)}_{n})(1-F(x^{(2)}_{n},x^{(1)}_{n}))|\leq \frac{\epsilon}{k}$\\\\
since  $f(x^{(2)}_{n})-f(x^{(1)}_{n}) < 0$, and  $x^{(1)}_{n}-x^{(2)}_{n} < 0$ (\ref{eq2}), then $x^{(1)}_{n}-\alpha(x^{(1)}_{n}-x^{(2)}_{n})(1-F(x^{(2)}_{n},x^{(1)}_{n}))\leq x^{(1)}_{n}+\frac{\epsilon}{k}$\\\\
Considering our target is $f(x_{n})-f(x^*) \leq \epsilon$ then we suppose $f(x_{n})-f(x^*) > \epsilon$\\\\
based on Lipschitz condition and (\ref{eq2}) we deduce that $f(x_{n})-f(x^*)\leq k (x^* - x_{n})\leq k (x^* - x^{(1)}_{n}) $\\\\  
So $x_{n+1}^{(1)}\leq x^{(1)}_{n}+\frac{\epsilon}{k}< x^*$\\\\
Therefore we got the result of the recurrence and we prove in the same manner the statement validity for the second sequence.\\\\

\textbf{Part 2:}\\\\
We proved in the first part that $x^{(1)}_{n}$ has an upper bound and $x^{(2)}_{n}$ has a lower bound, we still need to prove that $x^{(1)}_{n}$, $x^{(2)}_{n}$ converges toward the same limit to ensure they are adjacent sequences and from part 1 we deduce that this limit is $x^*$.\\\\
We get back to the definition of the sequences and since by construction one of them should change its value and if the other is stationary it will trivially be convergent, so we will focus on the non-stationary case:\\

 \begin{equation}
\quad \left\{
\begin{array}{l}
x^{(1)}_{n+1} = x^{(1)}_{n} -\alpha (x^{(1)}_{n}-x^{(2)}_{n})(1-F(x^{(2)}_{n},x^{(1)}_{n})), \text{ 
 when } f(x^{(2)}_{n})-f(x^{(1)}_{n})  < 0\\
x^{(2)}_{n+1} = x^{(2)}_{n} -\alpha (x^{(2)}_{n}-x^{(1)}_{n})(1+F(x^{(2)}_{n},x^{(1)}_{n})), \text{ 
 when } f(x^{(2)}_{n})-f(x^{(1)}_{n}) \geq 0
\end{array}
\right.
\end{equation}\\\\
We know that $\quad x^{(1)}_{n}-x^{(2)}_{n} < 0$, and on one hand for $f(x^{(2)}_{n})-f(x^{(1)}_{n})  < 0$, we have $\alpha (x^{(1)}_{n}-x^{(2)}_{n})(1-F(x^{(2)}_{n},x^{(1)}_{n})) < 0$, on the other hand for $f(x^{(2)}_{n})-f(x^{(1)}_{n})  \geq 0$, we have $\alpha (x^{(1)}_{n}-x^{(2)}_{n})(1-F(x^{(2)}_{n},x^{(1)}_{n})) > 0$. This means that $x^{(1)}_{n}$ is an increasing upper bounded sequence. ,$x^{(2)}_{n}$ is a decreasing lower bounded sequence therefore they converge respectively to the limits $l_1, l_2$.\\ 
Based on the expression of $x^{(1)}_{n+1}$ we get $\lim_{n \to \infty} x^{(1)}_{n+1} = \lim_{n \to \infty} x^{(1)}_{n} -\alpha (x^{(1)}_{n}-x^{(2)}_{n})(1-F(x^{(2)}_{n},x^{(1)}_{n})) = l_1 -\alpha (l_1-l_2)(1-F(l_2,l_1)) = l_1$.\\\\
Considering that $(1-F(x^{(2)}_{n},x^{(1)}_{n}))>0$ when $f(x^{(2)}_{n})-f(x^{(1)}_{n})  < 0$ and $\alpha >0$, then $l_2=l_1$\\
Similarly, we prove this statement in the second case.\\\\
From part 1, we have $x^* \leq l_1$,$l_2 \leq x^*$, and from part 2  $l_2=l_1$ we deduce that $l_1=l_2=x^*$ means that $x_n$ converges to $x^*$, which proves the validity of the theorem.

\end{proof}

\section{Results and Discussion}
In this section, we present our results from testing the Super Gradient Descent algorithm on various synthetic one-dimensional functions defined on a bounded interval and compare its performance to the classical optimization methods. The key metric considered is robustness in avoiding local minima and converging to global minimum.
\subsection{Gradient Descent vs Super Gradient Descent}
We first try to find the global minimum of the function \( f(x) = x \sin(x) \) which has several minima but it is enough regular.

\begin{figure}[H]
    \centering
    \begin{subfigure}[b]{0.8\textwidth}
        \centering
        \includegraphics[width=13cm, height=7cm]{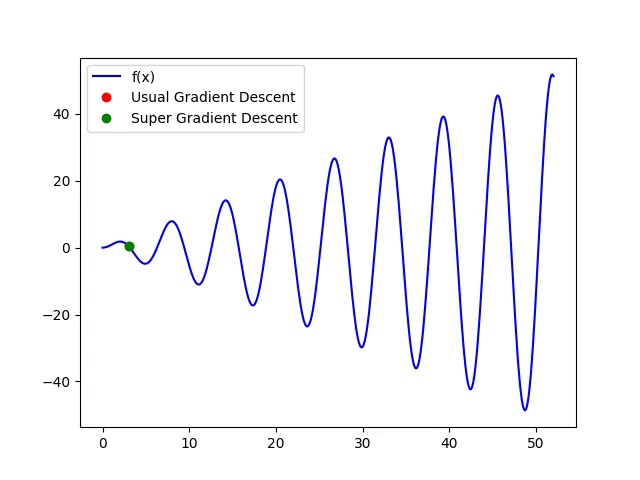}

        \label{fig:image3}
    \end{subfigure}

\vspace{-0.77cm} 

    \begin{subfigure}[b]{0.8\textwidth}
        \centering
        \includegraphics[width=13cm, height=7cm]{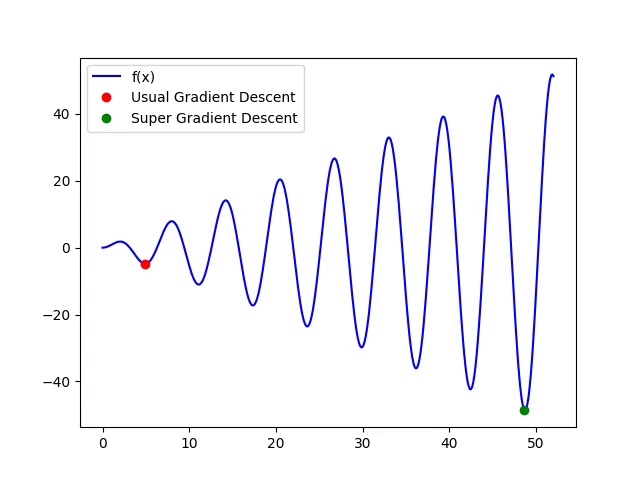}
  
        \label{fig:image6}
    \end{subfigure}

    \caption{Non-convex multi-minima test}
    \label{fig:combined1}
\end{figure}

The results of our experiments are illustrated in Figures \ref{fig:combined1}. As observed, the traditional gradient descent algorithm shows signs of convergence towards a local minimum, since the function admits several minima. The convergence path is often characterized by oscillations especially at final iterations, indicating a struggle to escape local minima.\\\\
In contrast, our proposed super gradient descent algorithm demonstrates a more stable and directed approach towards the global minimum. The algorithm effectively navigates the oscillations and maintains a consistent trajectory towards the global minimum. This enhanced performance is attributed to its global information-based update rules, which allow for more nuanced adjustments in the descent path.\\\\
Next, we apply both algorithms to the more complex function \( f(x) = 2x \sin(x^3) - x \cos\left(\frac{x^3}{12}\right) \) which presents not only the challenge of huge number of local minima but also the hard differentiability.

\begin{figure}[H]
    \centering
    \begin{subfigure}[b]{0.8\textwidth}
        \centering
        \includegraphics[width=13cm, height=7cm]{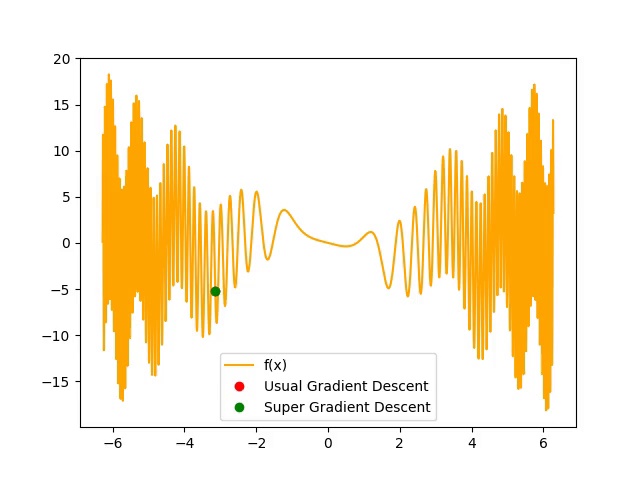}

        \label{fig:image3}
    \end{subfigure}

\vspace{-0.77cm} 

    \begin{subfigure}[b]{0.8\textwidth}
        \centering
        \includegraphics[width=13cm, height=7cm]{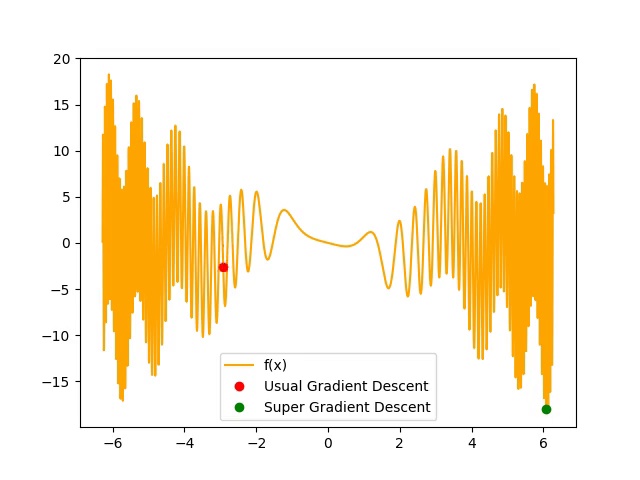}
  
        \label{fig:image6}
    \end{subfigure}

    \caption{Hardly differentiable function test}
    \label{fig:combined2}
\end{figure}

Here, the classical gradient descent again not only struggles with local minima, failing to reach the global minimum efficiently but also, the trajectory appears erratic and unstable due to the local sudden change in the objective function value, further confirming its sensitivity.\\\\
In contrast, the super gradient descent algorithm exhibits a more robust behavior, with a clear and smooth convergence towards the global minimum, regardless of the function's complexity. This can be observed in the final convergence plots where the loss reduction is significantly more consistent compared to the classical method.

\subsection{Benchmarking with the existing optimization algorithms}

In this section, we compare the performance of our proposed Super Gradient Descent (SuGD) algorithm with several state-of-the-art optimization algorithms defined in the second section, additionally to the Gradient Descent (GD), we tested AdaGrad, RMSprop, AdamW, NAG and Adam. The comparison focuses on the ability to escape local minima, the rate of convergence, and the consistency of reaching the global minimum.

\begin{figure}[H]
    \centering
    \begin{subfigure}[b]{0.8\textwidth}
        \centering
        \includegraphics[width=13cm, height=7cm]{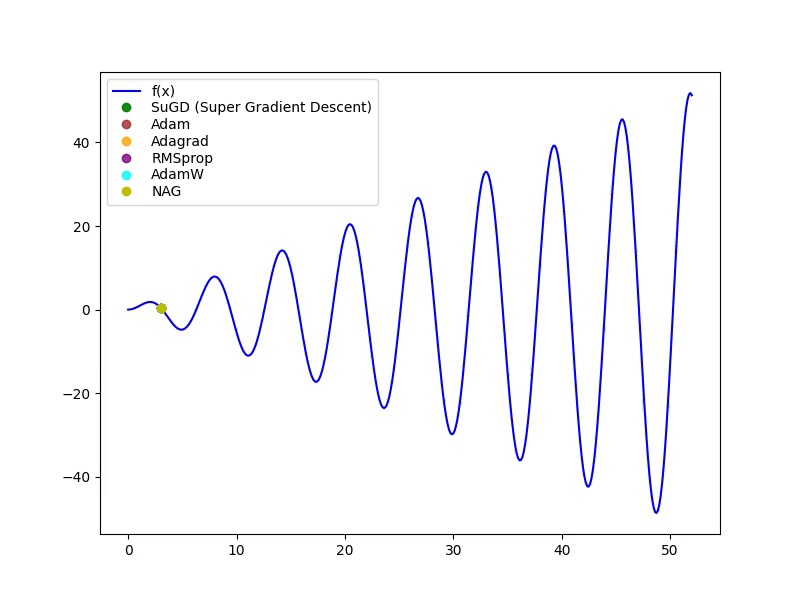}

        \label{fig:image3}
    \end{subfigure}

\vspace{-0.77cm} 

    \begin{subfigure}[b]{0.8\textwidth}
        \centering
        \includegraphics[width=13cm, height=7cm]{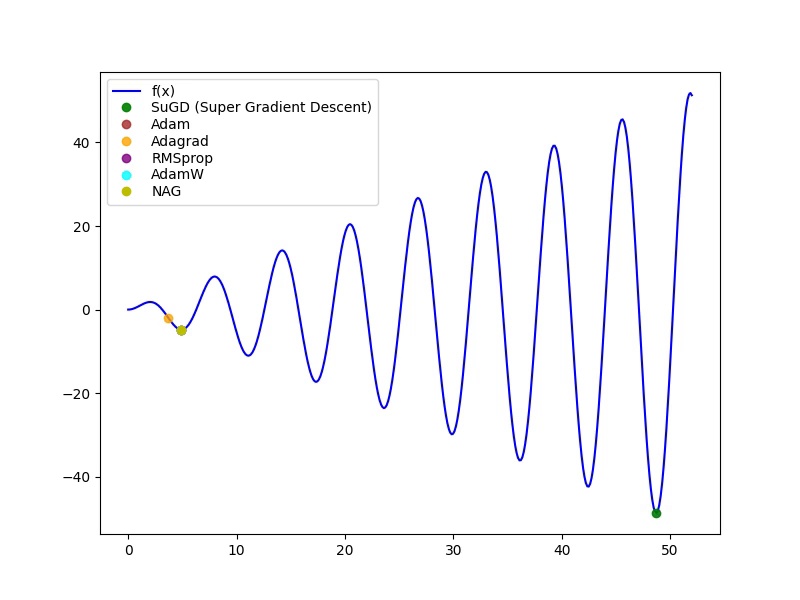}
  
        \label{fig:image6}
    \end{subfigure}

    \caption{Non-convex multi-minima test}
    \label{fig:combined3}
\end{figure}

In the first test function, 
$f(x) = x \sin(x)$, AdaGrad, which adapts its learning rate based on previous gradients \cite{duchi2011adaptive}, performs well at first but suffers from a decreasing learning rate over time, which causes premature convergence. RMSprop shows a more stable convergence but still fail to reach the global minimum in this case. The moment/gradient adaptive learning rates in AdamW and Adam accelerated the convergence but still not enough to escape the local minimum. The anticipating behaviour of NAG allowed it to climb in the ascending direction a bit but fastly got back to the local minimum.\\ 
In contrast, our proposed Super Gradient Descent effectively bypasses local minima and demonstrates a consistent path towards the global minimum and this within a reasonable time compared to the previous algorithms.

\begin{figure}[H]
    \centering
    \begin{subfigure}[b]{0.8\textwidth}
        \centering
        \includegraphics[width=13cm, height=7cm]{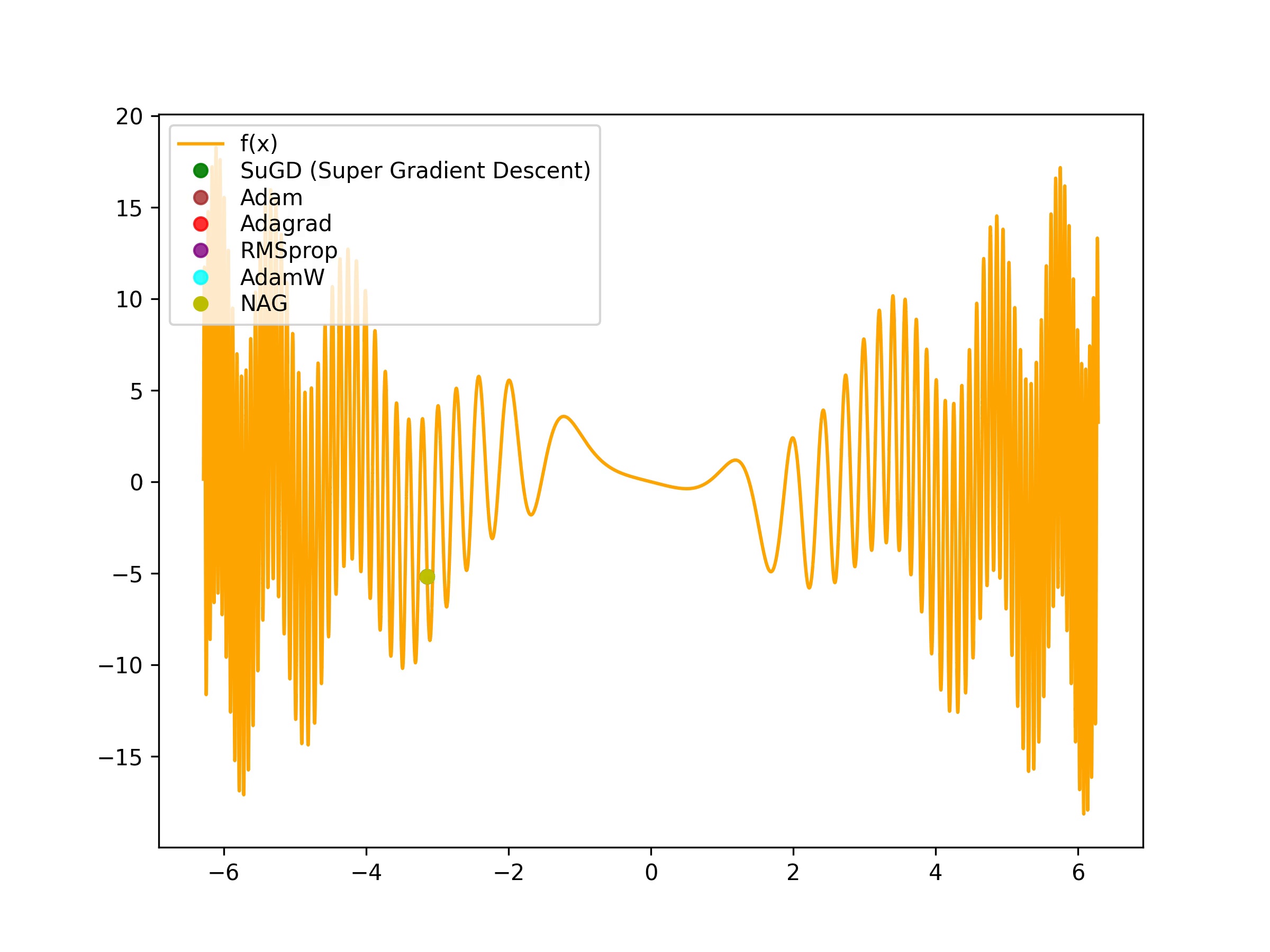}

        \label{fig:image3}
    \end{subfigure}

\vspace{-0.77cm} 

    \begin{subfigure}[b]{0.8\textwidth}
        \centering
        \includegraphics[width=13cm, height=7cm]{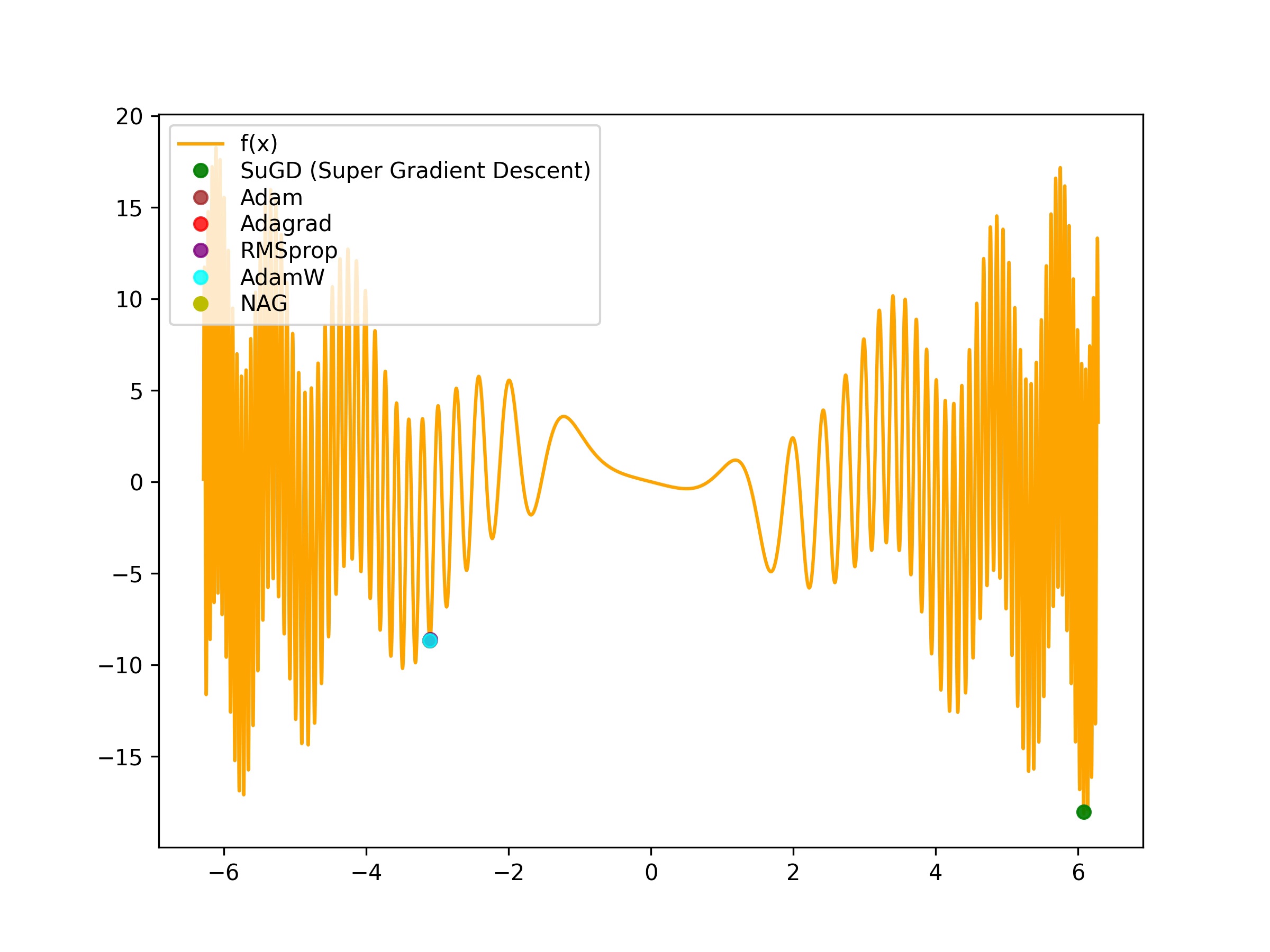}
  
        \label{fig:image6}
    \end{subfigure}

    \caption{Hardly differentiable non-convex function test}
    \label{fig:combined4}
\end{figure}

For the more complex function 
$f(x) = 2x \sin(x^3) - x \cos\left(\frac{x^3}{12}\right)$, the limitations of classical optimization algorithms become more apparent. NAG, while managing to escape local minima and reduce the loss, experienced a gradient exploding phenomenon and diverged because the function is much more complex and hardly differentiable. This highlights the stochastic behaviour in escaping the local minima of these algorithms and, therefore the unguarantee of the global convergence. In comparison, Super Gradient Descent displays much smoother convergence and consistently reaches the global minimum, avoiding the pitfalls of local minima encountered by other algorithms. The global information-based updates in our method allow it to adapt more effectively to the varying gradients and oscillations present in the function, guaranteeing the global convergence.\\\\
\begin{figure}[H]
    \centering
    \begin{subfigure}[b]{0.8\textwidth}
        \centering
        \includegraphics[width=13cm, height=8cm]{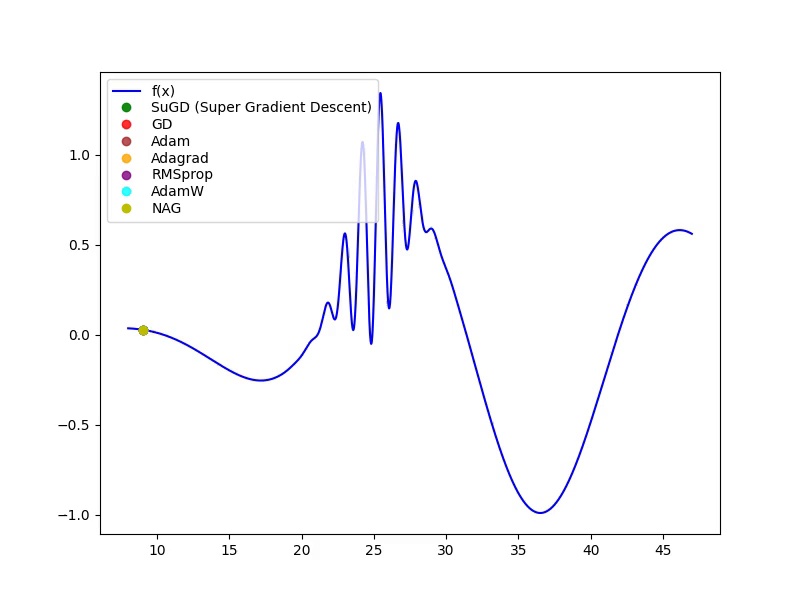}

        \label{fig:image3}
    \end{subfigure}

\vspace{-0.89cm} 

    \begin{subfigure}[b]{0.8\textwidth}
        \centering
        \includegraphics[width=13cm, height=8cm]{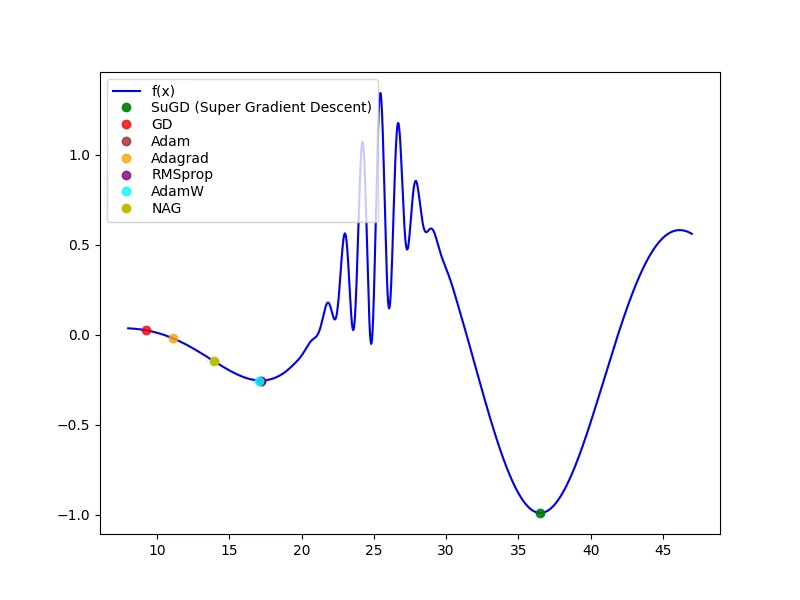}
  
        \label{fig:image6}
    \end{subfigure}

    \caption{Non-convex multi-regularity function test}
    \label{fig:combined5}
\end{figure}

In Figure \ref{fig:combined5} the experiment tests the Super Gradient Descent on a multi-regularity non-convex function defined as $f(x) = e^{-0.004(x-35)^2} \left( \sin(0.3x) + e^{-0.2(x-25)^2} \sin(5x) \right)
$. This function has two regular regions separated by a hardly differentiable region and the challenge here is not only escaping the local minimum but also being robust against the sudden change in regularity to achieve the global minimum region.\\\\
Overall, the results indicate that the super gradient descent algorithm not only converges to the global minimum effectively but also does so with greater stability and robustness than the traditional gradient descent method. This has significant implications for optimization problems in various fields, where the risk of being trapped in local minima can severely impact the performance of learning algorithms.\\\\
The comparative analysis presented here establishes the effectiveness of our proposed approach. Future work will involve testing on higher-dimensional functions and integrating additional optimization strategies to further enhance the robustness of our algorithm.

\section{Conclusion}
In this article, We presented the Super Gradient Descent novel algorithm capable of converging to the global minimum of any k-lipschitz one-dimensional
function defined on a domain $[a, b]$. We started by an overview about the state of the art optimization algorithms including Gradient Descent, AdaGrad, RMSprop, Adam, AdamW, Nesterov Accelerated Gradient (NAG). Then, we introduced our methodology and the mathematical implementation of the algorithm followed by the proof of the guaranteed global convergences showing how SuGD leverages global information about the function, more precisely the introduction of the concept of global gradient which plays a pivotal role in pointing to the global minimum. Finally we compared the performance of our algorithm with the existing optimization algorithms demonstrating its stable convergence and robustness in achieving the global minimum where the others struggle to escape local minima and maintain a consistent convergence path.\\ 
Overall, this novel algorithm succeeded in resolving one of the most challenging problems in optimization. Particularly, the guaranteed one-dimensional global convergence, on one hand it directly impacts the higher dimensional optimization through the line search method, and on the other hand, it opens up the perspectives for future expansion of this algorithm to a higher dimensional space.

\newpage

\end{document}